\documentclass{article}

\usepackage{iclr2027_conference,times}

\usepackage{amsmath,amsfonts,bm}

\def\eqref#1{equation~\ref{#1}}

\def\1{\bm{1}}

\DeclareMathAlphabet{\mathsfit}{\encodingdefault}{\sfdefault}{m}{sl}
\SetMathAlphabet{\mathsfit}{bold}{\encodingdefault}{\sfdefault}{bx}{n}

\usepackage{booktabs}
\usepackage{graphicx}
\usepackage{mathtools}
\usepackage{microtype}
\usepackage{multirow}
\usepackage{xcolor}
\usepackage{hyperref}
\usepackage{url}
\usepackage{capt-of}
\usepackage{comment}
\usepackage{enumitem}

\graphicspath{{figures/}}

\title{E8-QAT: Efficient 2-Bit Quantization-Aware Training with Implicit Lattice Codebooks}

\author{
  \begin{tabular}{@{}c@{}}
    Xinyu Lei\textsuperscript{1} \quad
    Hong Huang\textsuperscript{2} \quad
    Jiebin Zhang\textsuperscript{1} \quad
    Hongyu Li\textsuperscript{2} \quad
    Jinlong Song\textsuperscript{2} \\[0.5ex]
    Decheng Wu\textsuperscript{3} \quad
    Guanghua Yu\textsuperscript{3} \quad
    Jianchen Zhu\textsuperscript{3} \quad
    Sujian Li\textsuperscript{1} \\[1mm]
    {\small \textnormal{\textsuperscript{1}Peking University \quad
            \textsuperscript{2}City University of Hong Kong \quad
            \textsuperscript{3}Tencent}} \\[1mm]
    {\small \textnormal{\texttt{xinyulei25@stu.pku.edu.cn} \quad
            \texttt{lisujian@pku.edu.cn} \quad
            \texttt{lucayu@tencent.com}}}
  \end{tabular}
}

\iclrfinalcopy  

\begin{document}

\maketitle

\begin{abstract}
Low-bit quantization reduces the memory footprint of large language models, but preserving accuracy with efficient training remains challenging.
Scalar quantization encodes weights independently, while explicit-codebook vector quantization introduces costly search during training. We propose E8-QAT, a
2-bit quantization-aware training framework that uses the $E_8$ lattice as an implicit
eight-dimensional codebook. It jointly quantizes weight groups through arithmetic
projection without storing or searching an explicit codebook during the training forward
pass. A 2-bit core encoding and sparse extension preserve the selected lattice
points exactly, adding less than 0.1\% to core index storage. 
On Qwen3.6-35B-A3B, E8-QAT matches the performance of the full-precision model in terms of the average score across the four tasks.
E8-QAT improves over the state-of-the-art baseline by 2.03\% on IFEval and by 1.51\% on LiveCodeBench, under the same data, schedule, and bit width.
The 2-bit representation reduces peak memory by 80\% and supports a $4\times$ larger batch.
At each representation's largest feasible batch, E8-QAT reaches $1.97\times$ the decoding throughput of BF16.

\end{abstract}

\section{Introduction}\label{sec:introduction}

Large language models (LLMs) have made substantial progress across language tasks~\citep{wu2023overview}, but growing demands for privacy-preserving, offline, and cost-efficient deployment have made running them on resource-constrained platforms a pressing need~\citep{yao2024security,liagkou2024cost}. Weight quantization addresses this by reducing model size~\citep{awq}. Post-training quantization (PTQ) incurs low computational cost but can suffer accuracy degradation at ultra-low bit widths~\citep{gptq}. Quantization-aware training (QAT) recovers accuracy by allowing the model to adapt to quantization errors through weight updates, but training requires substantial data, compute, and memory~\citep{jacob2018integer,efficientqat}. This motivates quantization schemes that achieve strong accuracy at ultra-low bit widths without prohibitive training overhead.

Existing QAT approaches primarily rely on scalar quantization, which assigns quantization levels to each weight coordinate independently~\citep{paretoq}. While scalar quantization supports efficient per-coordinate arithmetic, it limits the flexibility of codeword arrangements across dimensions. Vector quantization (VQ) addresses this by jointly encoding groups of weights through a shared codebook~\citep{gray1984vector}, allowing richer codeword arrangements that better capture cross-dimensional weight structure and reduce quantization distortion at a given bit budget~\citep{aqlm,quipsharp}. Recent work has therefore explored integrating vector quantization into quantization-aware training (VQ-QAT)~\citep{pvtuning,lcqat}.

However, conventional VQ-QAT requires an explicit codebook that must be \textbf{stored and searched at every training step}, posing two challenges at scale: (1) nearest-codeword search during the training forward pass introduces substantial overhead, and (2) codebook size grows exponentially with vector dimension at a fixed bit width, making high-dimensional VQ impractical for both training and inference.

\begin{figure*}[t]
    \centering
    \includegraphics[width=\textwidth]{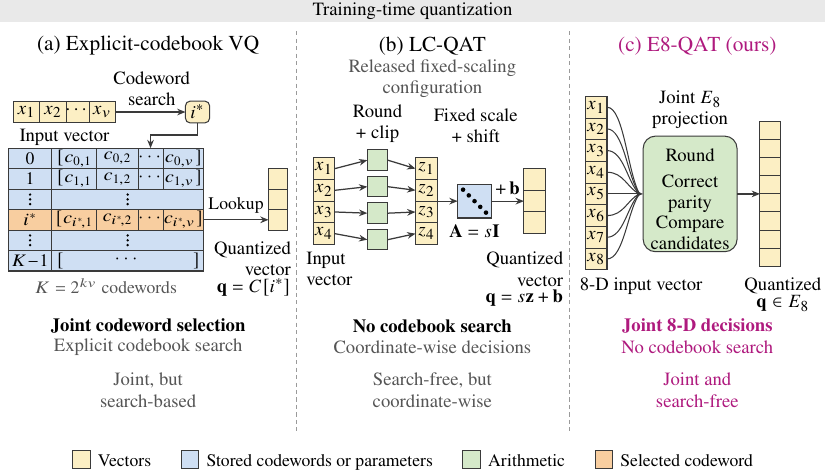}
    \caption{Training-time quantization.
(a) Explicit-codebook vector quantization searches a stored codebook of $K=2^{kv}$ entries.
(b) LC-QAT rounds each coordinate and applies a fixed scale and shift, $\mathbf{q}=s\mathbf{z}+\mathbf{b}$.
(c) E8-QAT projects eight normalized coordinates onto $E_8$ by arithmetic, without a stored codebook.}
    \label{fig:intro_quantization_comparison}
\end{figure*}

To address these challenges, we propose \textbf{E8-QAT}, a QAT framework that uses the $E_8$ lattice~\citep{conway1982fast} as an implicit eight-dimensional codebook, as illustrated in Figure~\ref{fig:intro_quantization_comparison}. During the training forward pass, each eight-dimensional weight group is projected onto the $E_8$ lattice by rounding, parity correction, and a constant number of candidate comparisons, 
reducing the per-group computational cost from $O(v\,2^{kv})$ for exhaustive codebook search to $O(v)$, independent of the nominal codebook size, where $k$ and $v$ denote the bit width and vector dimension, respectively.
Exploiting the observed concentration of quantized weight groups within the finite core, E8-QAT combines a 2-bit core encoding with a sparse extension that, at very low additional memory cost, exactly preserves the lattice points selected during training with an identity straight-through estimator (STE)~\citep{bengioste}.

Our main contributions are as follows:
\begin{itemize}[leftmargin=1.4em,itemsep=1pt,topsep=2pt,parsep=0pt]
\item \textbf{Search-free joint quantization for QAT.}
We jointly quantize each group of eight weights by closed-form projection
onto the infinite $E_8$ lattice in the QAT forward pass. The projection
requires $O(v)$ operations per group, independent of the nominal codebook
size, without codebook storage, search, or learning.

\item \textbf{Exact export beyond the finite core.}
Lattice projections during QAT can fall outside the finite core, and
clipping these points at export would alter the trained quantized weights.
We combine a dense core index with sparse extension records to preserve
every trained lattice point exactly. 

At the end of training, the sparse records add less than $0.1\%$ to the 2-bit index payload.

\item \textbf{Matched-budget 2-bit evaluation on agentic and long-form tasks.}
We evaluate the performance and efficiency of E8-QAT on Qwen3.6-35B-A3B~\citep{qwen36_35b_a3b} and Llama-3.1-8B~\citep{llama3}.

On Qwen3.6-35B-A3B, E8-QAT improves over LC-QAT by 2.03\% on IFEval and by 1.51\% on LiveCodeBench, under the same training corpus, schedule, and bit budget.
Its four-task average is 73.93, equal to the reported BF16 average.
Relative to BF16, the exported model reduces peak memory usage by 80\%,
enabling a $4\times$ larger batch size. Decoding throughput reaches
$1.97\times$ that of BF16 when each representation is evaluated at its
largest feasible batch size.
\end{itemize}
\section{Related Work}
\label{sec:related}

\subsection{Low-bit weight quantization}

Representative PTQ methods include GPTQ~\citep{gptq} and AWQ~\citep{awq}, which use calibration data to guide scalar quantization of individual weight coordinates.
Other PTQ methods use VQ to jointly encode groups of weights.
QuIP\#~\citep{quipsharp} uses predefined codebooks, while AQLM~\citep{aqlm} represents each weight group as a sum of codewords from learned codebooks.
QAT methods such as EfficientQAT~\citep{efficientqat} and ParetoQ~\citep{paretoq} incorporate scalar quantization into training, allowing model weights to adapt to quantization errors.

\subsection{Vector quantization-aware training}

Recent work has also explored VQ-QAT, with PV-Tuning and LC-QAT adopting different strategies for updating quantized weights.

PV-Tuning optimizes both continuous parameters and discrete codeword assignments.
For learned VQ codebooks, it combines gradient-based updates of codeword values with discrete assignment updates for selected weight groups.
These assignment updates require nearest-codeword search, or beam search when multiple additive codebooks are used, introducing additional computational cost during training~\citep{pvtuning}.

LC-QAT avoids explicit codebook lookup in the training forward pass through an affine mapping of discrete vectors~\citep{lcqat}.
However, it rounds and clips each coordinate independently before applying a fixed scale and shift~\citep{lcqatcode}.
The resulting quantizer retains the per-coordinate structure of scalar quantization, limiting the flexibility of codeword arrangements across dimensions.

Consequently, these VQ-QAT designs face a trade-off between training efficiency and quantization accuracy.
They either rely on costly codeword search to determine the discrete assignments, or give up the joint encoding that makes VQ accurate.
E8-QAT addresses this challenge by using the $E_8$ lattice as an implicit eight-dimensional codebook, enabling joint weight quantization through arithmetic projection~\citep{conway1982fast} without storing or searching an explicit codebook during the training forward pass.

\section{Preliminaries}
\label{sec:preliminaries}

At a fixed bit budget, quantization error depends on both the
distribution of the weights and the geometry of the codebook.
We first explain how Hadamard preprocessing redistributes weight
magnitudes, then introduce vector quantization as a way to
represent the transformed weights jointly. These considerations
motivate the use of the $E_8$ lattice, whose geometric properties
and efficient nearest-point projection provide the mathematical
basis for E8-QAT.

\subsection{Hadamard Preprocessing}
\label{sec:prelim_hadamard}

Randomized Hadamard preprocessing mixes weight coordinates before
quantization~\citep{quipsharp}. Within a transform block, it maps
a weight vector $\mathbf{w}$ to
\begin{equation}
    \widetilde{\mathbf{w}}=\mathbf{H}\mathbf{D}\mathbf{w},
    \qquad \mathbf{H}^{\top}\mathbf{H}=\mathbf{I},
    \label{eq:hadamard_transform}
\end{equation}
where $\mathbf{D}$ applies independent random sign flips and
$\mathbf{H}$ is a normalized Hadamard matrix. Each output
coordinate is a normalized sum of signed weights from the
same block, while the transform preserves the sum of
squared weights.

The central limit theorem~\citep{durrett2019probability} shows that when many independent signed weights contribute to each output coordinate and no single weight dominates, the resulting marginal distribution approximates a Gaussian distribution. To verify this, we sampled 14 weight tensors from a pre-quantized BF16 model and observed that the transformed marginal kurtosis is about $3.02$, close to the Gaussian value $3$.

Under the zero-mean isotropic Gaussian approximation, the probability density depends only on the distance from the origin and decreases with that distance. We observed that the radial probability distribution of the rotated weights is geometrically similar to a finite $E_8$ codebook~\citep{quipsharp}, which can improve lattice utilization and representation accuracy in vector quantization.

\subsection{Vector Quantization}
\label{sec:prelim_vq}

After Hadamard preprocessing and scaling, we partition each row
of a weight matrix into non-overlapping groups of $d$ consecutive
weights along the input dimension. Let
$\mathbf{x}\in\mathbb{R}^{d}$ denote one such group.
Given a finite codebook $\mathcal{C}\subset\mathbb{R}^{d}$
with $K$ codewords, vector quantization assigns the entire group
to its nearest codeword:
\begin{equation}
    Q_{\mathcal{C}}(\mathbf{x})
    = \operatorname*{arg\,min}_{\mathbf{c}\in\mathcal{C}}
      \|\mathbf{x}-\mathbf{c}\|_2^2.
    \label{eq:vq_assignment}
\end{equation}
With $K=2^{kd}$ codewords, each group requires a $kd$-bit index.

Unlike scalar quantization, VQ is not restricted to a rectangular
grid and can better match the distribution of weight groups.
This flexibility can reduce quantization error at the same
index budget~\citep{grayneuhoff1998}.

However, exhaustive nearest-codeword search costs $O(dK)$
operations per group, with $K=2^{kd}$ growing exponentially
in the group dimension at a fixed per-weight bit budget.
This motivates structured point sets whose nearest points
can be computed directly rather than found by exhaustive search.

\subsection{\texorpdfstring{$E_8$}{E8} Lattice Quantization}
\label{sec:prelim_e8}

The $E_8$ lattice is a classical eight-dimensional construction
from sphere-packing and lattice-quantization theory
\citep{conwaysloane1999}, not a codebook learned from model weights.
Its points have either all integer or all half-integer
coordinates, with an even coordinate sum:
\begin{equation}
    E_8 = \left\{
        \mathbf{z}\in\mathbb{Z}^{8}\cup
        \left(\mathbb{Z}^{8}+\tfrac{1}{2}\mathbf{1}\right)
        :\ \mathbf{1}^{\top}\mathbf{z}\in 2\mathbb{Z}
    \right\},
    \label{eq:e8_definition}
\end{equation}
where $\mathbf{1}$ is the all-ones vector. Its geometry and
algebraic structure offer three useful properties.

$E_8$ combines favorable covering geometry with an algebraic
structure that enables efficient exact quantization. At equal
cell volume, its smaller covering radius than $\mathbb{Z}^{8}$
provides a tighter worst-case error bound throughout the
infinite lattice~\citep{conwaysloane1999}, while its structure
permits exact nearest-point projection without exhaustive
codebook search~\citep{conway1982fast}. This combination makes
$E_8$ attractive for QAT, where weights are repeatedly projected
as they evolve during training. We exploit these properties
by using an approximately ball-shaped core
$\mathcal{C}_{\mathrm{core}}\subset E_8$ of $2^{16}$ low-norm
points for compact encoding, while retaining the full lattice
as the training-time projection space. This design pairs
$2$-bit-per-weight indices for core points with the flexibility
to project beyond the core boundary; the following section
describes how these core-external projections are encoded
and their additional cost is accounted for.

\section{Method}
\label{sec:method}

\begin{figure}[t]
    \centering
    \includegraphics[width=\linewidth]{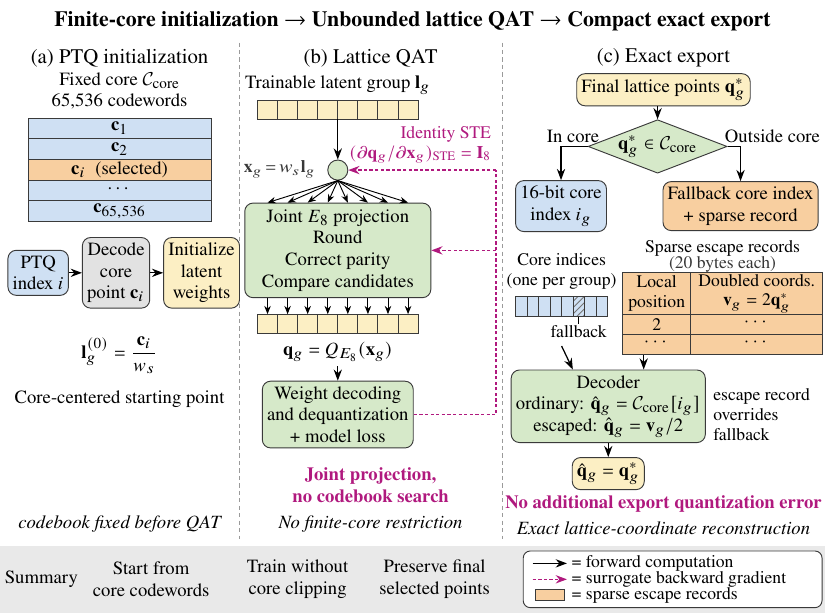}
    \caption{
    Overview of E8-QAT.
    (a) PTQ initializes latent weights from a fixed
    \(E_8\) core.
    (b) Arithmetic lattice projection with an identity STE
    enables joint QAT without explicit codebook search or
    finite-core clipping.
    (c) Core indices and sparse escape records preserve the
    final lattice points.
    Hadamard transforms and reconstruction scales are omitted
    for clarity.
    }
    \label{fig:method_overview}
\end{figure}

E8-QAT separates training-time quantization from post-training
encoding, combining efficient joint quantization with compact
deployment.
As illustrated in Figure~\ref{fig:method_overview}, the framework
starts from finite-core PTQ codewords, performs lattice QAT with
an identity straight-through estimator, and preserves the
final projections using core indices and sparse extensions.

\subsection{Lattice QAT from PTQ Initialization}
\label{sec:method_qat}
\label{sec:method_init}

Let \(\mathbf{L}\in\mathbb{R}^{m\times n}\) denote the trainable
latent weights of a layer with \(n\) input and \(m\) output
channels.
Assuming \(n\) is divisible by \(d=8\), we partition each row
into contiguous groups
\(\boldsymbol{\ell}_g\in\mathbb{R}^{d}\).
The quantization mapping is
\begin{equation}
    \mathbf{x}_g=w_s\boldsymbol{\ell}_g,
    \qquad
    \mathbf{q}_g=Q_{E_8}(\mathbf{x}_g),
    \qquad
    \overline{\mathbf{w}}_g
    =c_s\mathbf{q}_g+c_z\mathbf{1}_d,
    \label{eq:method_coordinates}
\end{equation}
where group \(g\) contains \(d=8\) latent weights \(\boldsymbol{\ell}_g\).
The fixed per-tensor scale \(w_s\) maps them to lattice coordinates \(\mathbf{x}_g\),
and \(Q_{E_8}\) returns the nearest lattice point \(\mathbf{q}_g\).
The fixed decode scale \(c_s\) and zero-point \(c_z\) map that point back to weights;
\(\mathbf{1}_d\) adds \(c_z\) to every coordinate.

\paragraph{Weight reconstruction.}
The decoded groups in Equation~\ref{eq:method_coordinates} are assembled
into $\overline{\mathbf{W}}\in\mathbb{R}^{m\times n}$. The effective weight
matrix is reconstructed as
\begin{equation}
    \widehat{\mathbf{W}}
    =
    \operatorname{diag}(\mathbf{s}_V)\,
    H_{\mathrm{out}}\,
    \overline{\mathbf{W}}\,
    H_{\mathrm{in}}\,
    \operatorname{diag}(\mathbf{s}_U),
    \label{eq:method_effective_weight}
\end{equation}
where $\mathbf{s}_U\in\mathbb{R}^{n}$ and
$\mathbf{s}_V\in\mathbb{R}^{m}$ are trainable input- and output-channel
scaling vectors, respectively. The matrices $H_{\mathrm{in}}$ and
$H_{\mathrm{out}}$ are fixed orthogonal block-Hadamard transforms
with block size $128$. For an input $\mathbf{a}\in\mathbb{R}^{n}$,
the layer computes $\mathbf{y}=\widehat{\mathbf{W}}\mathbf{a}$.

\paragraph{Finite-core construction.}
The finite core $\mathcal{C}_{\mathrm{core}}$ consists of the first
$2^{16}$ points of $E_8$ ordered by increasing squared norm, with ties
resolved lexicographically. It includes all $56{,}881$ points with
squared norm at most $10$ and the first $8{,}655$ points in lexicographic
order from the shell at squared norm $12$. Initialization, export,
and decoding use the same point set and index ordering.
Since the outermost shell is only partially retained, a norm threshold
does not suffice for escape detection; a projected point is classified
as an escape if and only if it lies outside $\mathcal{C}_{\mathrm{core}}$.

\paragraph{Numerical precision.}
Latent weights are stored in BF16, while projection candidates and squared distances are evaluated in FP32.
Gradients through the hard projection use an identity straight-through estimator.
Hadamard transforms, affine decoding, and channel scaling lie outside the projector and use ordinary backpropagation.

\paragraph{PTQ initialization.}
We initialize from an \(E_8\)-based PTQ checkpoint obtained
using Hadamard preprocessing and LDLQ~\citep{quipsharp,quip}.
PTQ uses a fixed core
\(\mathcal{C}_{\mathrm{core}}
=\{\mathbf{c}_i\}_{i=0}^{2^{16}-1}\subset E_8\),
shared with the export representation.
Given the PTQ index \(i_g^{(0)}\), we initialize
\begin{equation}
    \boldsymbol{\ell}_g^{(0)}
    =
    \frac{\mathbf{c}_{i_g^{(0)}}}{w_s}.
    \label{eq:method_ptq_init}
\end{equation}
Thus, \(\mathbf{x}_g^{(0)}=\mathbf{c}_{i_g^{(0)}}\) in exact
arithmetic.
This places each group at a quantization-cell center, providing
a local stability margin against small perturbations~\citep{conwaysloane1999}.
The finite-precision condition is given in
Appendix~\ref{app:init_impl}; subsequent optimization is not
constrained to remain within the core.

\paragraph{Lattice training.}
Each forward pass applies the arithmetic projector in
Equation~\ref{eq:method_coordinates}.
Rounding, parity correction, and comparison of two candidates
require \(O(d)\) operations per group~\citep{conway1982fast}, without explicit codebook
search or restriction to the finite core.
Since hard projection is piecewise constant, we use an identity
STE~\citep{bengioste,vqvae}.
For the autoregressive language-modeling loss \(\mathcal{L}\),
the forward computation and surrogate latent gradient are
\begin{equation}
    \begin{aligned}
        \mathbf{q}_g
        &=
        \mathbf{x}_g+
        \operatorname{sg}\!\left(
            Q_{E_8}(\mathbf{x}_g)-\mathbf{x}_g
        \right),  
        \frac{\partial\mathcal{L}}
             {\partial\boldsymbol{\ell}_g}
        &\approx
        w_s c_s
        \frac{\partial\mathcal{L}}
             {\partial\overline{\mathbf{w}}_g},
    \end{aligned}
    \label{eq:method_ste}
\end{equation}
where \(\operatorname{sg}\) is the identity in the forward pass
and has zero derivative in the backward pass.
This substitutes \(I_d\) for the projection Jacobian while
retaining the hard forward assignment.

The assignment therefore remains jointly constrained by
\(E_8\), while latent updates can move groups across Voronoi
boundaries.
We optimize the latent weights, per-channel scales, and other
trainable parameters without learning the lattice or its core
codewords.
The STE supplies a surrogate gradient rather than the true
derivative of hard projection.

\subsection{Exact Encoding with Sparse Extensions}
\label{sec:method_export}

Training on the infinite lattice permits projections outside
the finite core.
Let
\(\mathbf{q}_g^\star=Q_{E_8}(\mathbf{x}_g^{(T)})\)
denote the final projection after \(T\) training steps.
For \(G\) quantized groups, define
\begin{equation}
    \mathcal{E}
    =
    \left\{
        g:\mathbf{q}_g^\star
        \notin\mathcal{C}_{\mathrm{core}}
    \right\},
    \qquad
    \rho=\frac{|\mathcal{E}|}{G}.
    \label{eq:method_escape_rate}
\end{equation}
An escape is determined by core membership, not solely by
a norm threshold.
Reprojecting escaped points onto the core would alter the
quantized weights selected during training.

Instead, we store a 16-bit core index \(i_g\) for every group.
For an in-core group, it represents \(\mathbf{q}_g^\star\)
directly; for an escaped group, it identifies a legal fallback.
Each escape additionally stores its position and doubled
coordinates
\(\mathbf{v}_g=2\mathbf{q}_g^\star\in\mathbb{Z}^{d}\).
Decoding gives
\begin{equation}
    \widehat{\mathbf{q}}_g
    =
    \begin{cases}
        \mathbf{c}_{i_g},
            & g\notin\mathcal{E},\\[2pt]
        \mathbf{v}_g/2,
            & g\in\mathcal{E}.
    \end{cases}
    \label{eq:method_exact_decode}
\end{equation}
The sparse record overrides the fallback rather than adding
absolute coordinates to it.
When positions and doubled coordinates fit the representation,
\(\widehat{\mathbf{q}}_g=\mathbf{q}_g^\star\) for every group.
Export therefore introduces no additional lattice-coordinate
quantization error.

With \(B\) additional bytes per escape, the payload cost is
\begin{equation}
    b_{\mathrm{payload}}
    =
    \frac{16G+8B|\mathcal{E}|}{dG}
    =
    2+\frac{8B}{d}\rho
    \quad\text{bits per quantized weight}.
    \label{eq:method_payload}
\end{equation}
Our coordinate format uses \(B=20\), giving
\(b_{\mathrm{payload}}=2+20\rho\) for \(d=8\).
This accounts for core indices and escape records, excluding
shared codebooks, scales, and format metadata.
Appendix~\ref{app:sparse_encoding} details the representation,
range checks, and optional root packing.
Section~\ref{sec:encoding_overhead} measures the encoding
overhead empirically.

After export, latent weights are discarded, leaving core
indices, sparse records, and reconstruction parameters.
Inference may use a shared decoding table, whereas
training-time projection remains lookup-free.
\section{Experiments}
\label{sec:experiments}

\subsection{Experimental Setup}
\label{sec:exp:setup}

\begin{table}[t]
\centering
\footnotesize
\setlength{\tabcolsep}{5.5pt}
\caption{Results at 2 bits on Qwen3.6-35B-A3B. Bold marks the best quantized result in
each column. Full results, including ARC-Easy, BoolQ, 
are given in Appendix~\ref{app:full}.}
\label{tab:exp:main}
\begin{tabular}{llccccccccc}
\toprule
& & & \multicolumn{5}{c}{Base (lm-eval)} & \multicolumn{2}{c}{Agent} & IF \\
\cmidrule(lr){4-8}\cmidrule(lr){9-10}\cmidrule(lr){11-11}
Type & Method & Bits & ARC-C & HelS & PIQA & WinG & Avg. & LCB & BFCL & IFEval \\
\midrule
N/A & BF16 & 16 & 
55.20
& 83.49 & 82.75 & 74.27 & 73.93 & 81.42 & 75.99 & 85.40 \\
\midrule
\multirow{3}{*}{SQ}
& RTN    & 2 & 49.23 & 79.44 & 79.87 & 69.85 & 69.60 & 20.38 & 53.91 & 77.82 \\
& GPTQ   & 2 & 52.78 & 79.17 & 80.20 & 69.76 & 70.48 & 20.38 & 55.87 & 78.18 \\
& UniSVQ & 2 & 56.83 & 78.31 & 80.58 & 71.59 & 71.83 & 31.28 & 72.32 & 80.41 \\
\midrule
\multirow{4}{*}{VQ}
& QuIP\# & 2 & 56.66 & 77.25 & 80.20 & 72.77 & 71.72 & 47.77 & 67.31 & 81.70 \\
& QTIP   & 2 & 51.71 & 79.36 & 81.12 & 72.69 & 71.22 & 26.45 & 61.77 & 80.78 \\
& AQLM   & 2 & 54.95 & 79.59 & 81.50 & 73.32 & 72.34 & 52.32 & 75.14 & 81.52 \\
& E8-PTQ & 2 & 56.74 & 79.38 & 80.47 & 72.61 & 72.30 & 55.73 & \textbf{76.14} & 81.89 \\
\midrule
\multirow{3}{*}{VQ-QAT}
& PV-Tuning & 2 & 56.57 & 80.89 & \textbf{82.15} & 73.48 & 73.27 & 75.73 & 74.83 & 81.89 \\
& LC-QAT & 2 & 56.57 & 81.28 & 80.74 & 73.88 & 73.12 & 78.58 & 75.22 & 80.96 \\
& E8-QAT & 2 & \textbf{57.85} & \textbf{82.29} & 81.56 & \textbf{74.03} & \textbf{73.93}
         & \textbf{80.09} & 74.16 & \textbf{82.99} \\
\bottomrule
\end{tabular}

\end{table}

\begin{table}[t]
\begin{minipage}[t]{0.56\textwidth}
\vspace{0pt}
\centering
\label{tab:exp:llama}
\footnotesize
\setlength{\tabcolsep}{3pt}
\begin{tabular}{@{}llccccc@{}}
\toprule
& & \multicolumn{5}{c}{Base (lm-eval)} \\
\cmidrule(l){3-7}
Type & Method & ARC-C & HelS & PIQA & WinG & Avg. \\
\midrule
N/A & BF16 & 55.55 & 79.55 & 81.28 & 73.80 & 72.54 \\
\midrule
\multirow{3}{*}{SQ}
& RTN  & 25.34 & 26.54 & 51.90 & 51.70 & 38.87 \\
& GPTQ & 24.40 & 26.38 & 50.98 & 50.04 & 37.95 \\
& UniSVQ & 39.51 & 63.55 & 72.63 & 66.46 & 60.54 \\
\midrule
\multirow{2}{*}{VQ}
& E8-PTQ & 43.94 & 66.16 & 74.76 & 68.90 & 63.44 \\
\midrule
\multirow{2}{*}{VQ-QAT}
& LC-QAT & 40.10 & 67.71 & 73.29 & 67.40 & 62.12 \\
& E8-QAT & \textbf{44.62} & \textbf{70.03} & \textbf{76.17} & \textbf{69.61} & \textbf{65.11} \\
\bottomrule
\end{tabular}
\caption{Results at 2 bits on Llama-3.1-8B. Bold marks the best quantized result in each column. Full results, including ARC-Easy, BoolQ, and WikiText perplexity, are given in Appendix~\ref{app:llama}.}
\end{minipage}\hfill
\begin{minipage}[t]{0.40\textwidth}
\vspace{0pt}
\centering
\includegraphics[width=\linewidth]{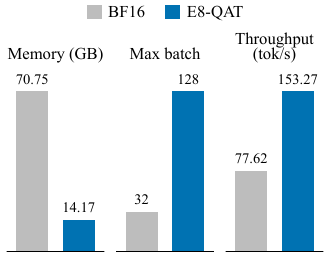}
\captionof{figure}{Serving capacity. Memory is measured at batch 1 and throughput at
each model's largest feasible batch.}
\label{fig:exp:infer}
\end{minipage}
\end{table}

We summarize the experimental configuration below and provide further details in Appendix~\ref{app:exp_protocol}.

\paragraph{Baselines.}
We compare E8-QAT against two categories of 2-bit methods:
(1) post-training quantization, including round-to-nearest (RTN) and GPTQ;UniSVQ~\citep{unisvq,lcqatcode}, the scalar PTQ used to initialize LC-QAT;
AQLM with a single learned $2^{16}$-entry codebook shared across all feed-forward
matrices of a layer, which matches our bit width; and E8-PTQ, the first stage of our method; and
(2) quantization-aware training, including PV-Tuning, initialized from the same
AQLM model, and LC-QAT.

\paragraph{Models, data, and evaluation.}
We use Qwen3.6-35B-A3B and Llama-3.1-8B as our base models.
Both stages draw on the UltraData collection~\citep{ultradata}: stage-1 calibration uses 1,680 sequences of 16,384 tokens constructed from 2,000 agentic conversations in UltraData-SFT-Agent, shared by all PTQ methods, and stage-2 training uses 2,000 long sequences from the same collection, totaling 50.8M tokens per pass, with a 375-step schedule for Llama-3.1-8B.
We evaluate model performance with lm-evaluation-harness~\citep{lmeval} on four zero-shot benchmarks, ARC-Challenge (ARC-C)~\citep{arc}, HellaSwag (HelS)~\citep{hellaswag}, PIQA~\citep{piqa}, and WinoGrande (WinG)~\citep{winogrande}, together with LiveCodeBench (LCB)~\citep{livecodebench} pass@1, BFCL~\citep{bfcl} v3 overall accuracy, IFEval prompt-level strict accuracy~\citep{ifeval}, and GPQA~\citep{gpqa} accuracy.
All models are evaluated with the same harness and settings; details are given in Appendix~\ref{app:exp_protocol}.

\begin{figure}[t]
\centering
\includegraphics[width=\linewidth]{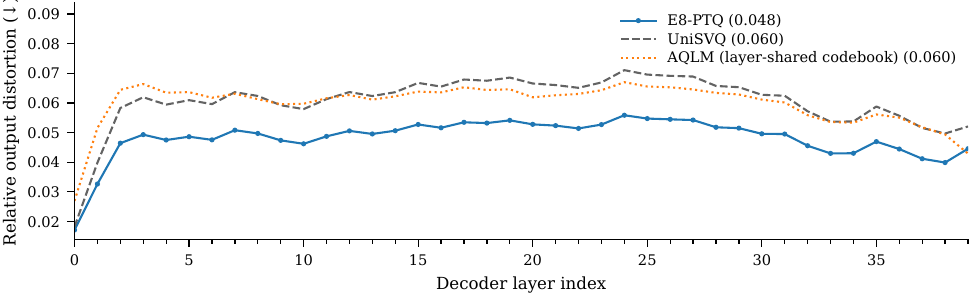}
\caption{Layer-wise stage-1 output distortion at matched bit width (lower is better).}
\label{fig:exp:stage1}
\end{figure}

\paragraph{Implementation details.}
All Qwen3.6-35B-A3B training runs use eight GPUs.
E8-QAT and LC-QAT use 750 steps, a warmup--stable--decay schedule annealed to zero,
and a peak learning rate of $10^{-4}$.
The Llama-3.1-8B second stage uses 375 steps, with the peak learning rate scaled by the inverse weight scale.
PV-Tuning uses its released recipe. In paired 375-step runs, the E8-QAT stage completes 4.5\% faster than LC-QAT.
Inference is measured with fused decode kernels and 16-bit indices, launching a separate process for each configuration.

\subsection{Main Results}
\label{sec:exp:main}

\paragraph{Comparison with 2-bit methods.}
To evaluate whether the lattice codebook yields the strongest 2-bit model at a matched bit budget,
we compare E8-QAT against the PTQ and QAT baselines on Qwen3.6-35B-A3B.
As shown in Table~\ref{tab:exp:main}, E8-QAT achieves the best quantized score on ARC-Challenge, HellaSwag, and WinoGrande, and on LiveCodeBench and IFEval.
Compared with LC-QAT, which shares the same data, schedule, and bit width, E8-QAT improves accuracy by 1.51 percentage points on LiveCodeBench and by 2.03 percentage points on IFEval.

On the six-task average, E8-QAT scores 76.95, against 75.97 for LC-QAT.
On Llama-3.1-8B, its WikiText perplexity is 13.77, against 16.73 for LC-QAT.Appendix~\ref{app:llama} gives ARC-Easy, BoolQ, and the perplexity for every Llama row.
These results demonstrate that replacing the scalar grid with the $E_8$ lattice improves both base-model quality and agentic capability at 2 bits.

\paragraph{Generalization to dense models.}
To examine whether this advantage transfers across architectures, we further evaluate E8-QAT on
the dense Llama-3.1-8B.
As shown in Table~\ref{tab:exp:llama}, E8-QAT again achieves the best quantized score on all four
tasks, outperforming LC-QAT by 2.99\% on average.
These results confirm that the lattice advantage transfers across architectures once the learning
rate is matched in lattice space.

\paragraph{Inference efficiency.}
We compare the serving capacity of E8-QAT and BF16 to validate that 2-bit storage allows a single GPU to serve larger batches
As shown in Figure~\ref{fig:exp:infer}, E8-QAT reduces peak memory by 80\% relative to BF16,
supports a $4\times$ larger batch, and achieves 97\% higher throughput at each model's largest feasible batch.
These results indicate that the lattice model nearly doubles serving throughput on the same hardware.

\paragraph{Most accurate initialization.}
At equal bits, the lattice should reconstruct weights more accurately than a scalar grid or
a learned codebook.
We measure the layer-wise output distortion
, which weighted by
routed-token frequency, with the same Hessians for all methods.
As shown in Figure~\ref{fig:exp:stage1}, E8-PTQ has the lowest distortion in 39 of the
40 layers.
Averaged over layers, it is 20.4\% below UniSVQ, which runs the same LDLQ pipeline with a
scalar grid, and 19.4\% below AQLM.
These results confirm that the lattice provides the most accurate starting point at matched bits.

\begin{figure}
\centering
\includegraphics[width=\linewidth]{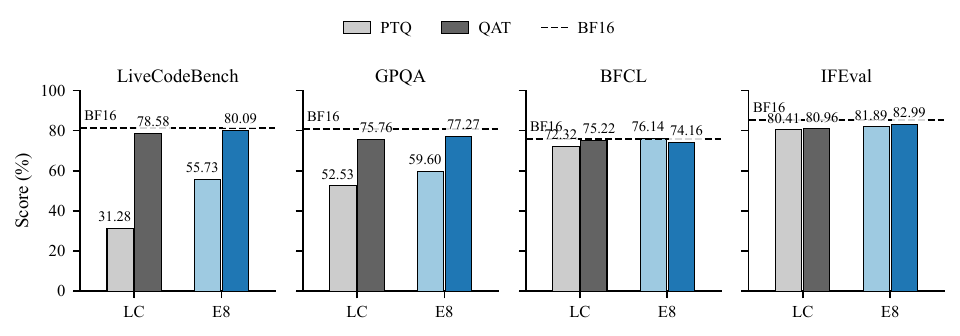}
\caption{Performance of PTQ and QAT model with the 750-step second stage.
Dashed lines mark BF16.}
\label{fig:exp:qat-gain}
\end{figure}

\paragraph{Gains from second-stage QAT.}
Reconstruction-based PTQ is expected to preserve short-form ability but not long-form generation,
which QAT should recover.
To verify this, Figure~\ref{fig:exp:qat-gain} compares each pipeline before and after the 750-step second stage.
QAT improves LiveCodeBench by 24.36\% and GPQA by 17.67\% for E8-QAT.
E8-QAT is ahead of the LC-QAT pipeline both before and after this stage. 
These results demonstrate that second-stage QAT is necessary for long-form generation at 2 bits and
that the lattice advantage persists through training.

\begin{figure}[t]
\centering
\includegraphics[width=\linewidth]{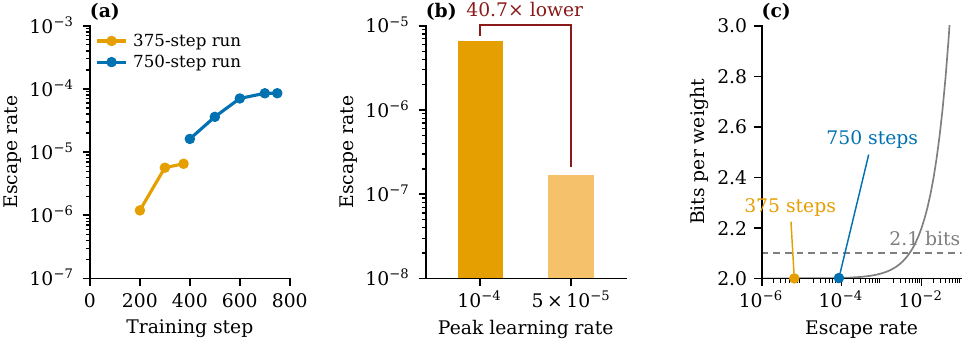}
\caption{Escape trajectories (a), peak-learning-rate control at 375 steps (b), and the
index-and-extension payload $b = 2 + 20\rho$ (c).}
\label{fig:exp:escape}
\end{figure}

\paragraph{Exact export at two bits.}
\label{sec:encoding_overhead}
Allowing QAT to move groups outside the finite core is expected to incur little additional storage.
To quantify this overhead, Figure~\ref{fig:exp:escape} tracks the escape rate during training, its
sensitivity to the peak learning rate, and the resulting payload.
Escape growth slows as the learning rate anneals, with the final 50 steps of the 750-step run raising
the escape rate by only 0.25\%; halving the peak learning rate reduces the final escape rate by
$40.7\times$; and at the 750-step endpoint, the sparse records add only 0.086\% to the 2-bit index payload.
These results confirm that a finite core with a sparse extension keeps exact export within 0.1\% of
two bits, preserving the trained weights exactly with only a small number of exceptions.

\section{Conclusion}
\label{sec:conclusion}

In this work, we introduce E8-QAT, a QAT framework designed to address the trade-off between training efficiency and quantization accuracy at ultra-low bit widths.
By using the $E_8$ lattice as an implicit eight-dimensional codebook, E8-QAT enables joint weight quantization through arithmetic projection without storing or searching an explicit codebook during the training forward pass.
An identity STE supports weight updates during training.
To preserve the lattice points selected during training, we combine a 2-bit core encoding with a sparse extension, enabling exact encoding at very low additional memory cost in our experiments.

Experimental results on Qwen3.6-35B-A3B and Llama-3.1-8B show that E8-QAT achieves higher average accuracy than LC-QAT
on the four zero-shot benchmarks. 
On Qwen3.6-35B-A3B, E8-QAT also reduces peak memory by 80\% relative to BF16 at batch size 1 and achieves $1.97\times$ the BF16 serving throughput on the same hardware, with throughput measured at each model's largest feasible batch.

\newpage
\subsubsection*{AI Use Statement}
Generative AI tools were used to aid and polish the writing.
This included language editing, translation, \LaTeX{} formatting, and drafting the presentation of figures and tables.
The authors reviewed all AI-assisted text and independently checked the mathematical derivations, experimental claims, citations, and generated artifacts.
The authors take responsibility for the final content.

\bibliography{references}
\bibliographystyle{iclr2027_conference}

\appendix
\section{Initialization and Implementation Details}
\label{app:init_impl}

\paragraph{Geometric margin over the integer lattice.}
At equal cell volume, \(E_8\) has covering radius smaller than \(\mathbb{Z}^{8}\) by a factor of \(\sqrt{2}\), and its normalized second moment is \(0.0716821\) against \(1/12\) for \(\mathbb{Z}^{8}\) \citep{conwaysloane1999}.
This is a granular gain of \(0.654\)\,dB, or a \(14.0\%\) reduction in mean squared error at matched rate under a high-resolution model.

\paragraph{Local stability of initialization.}
In the standard normalization of \(E_8\), the minimum distance between distinct \(E_8\) points is \(\sqrt{2}\).
Let
\[
    \epsilon_g^{(0)}
    =
    \left\|
        \mathbf{x}_g^{(0)}
        -\mathbf{c}_{i_g^{(0)}}
    \right\|_2
\]
denote the finite-precision initialization error of group \(g\).
The initial assignment is preserved at step \(t\) whenever
\begin{equation}
    \epsilon_g^{(0)}
    +
    \left\|
        \mathbf{x}_g^{(t)}-\mathbf{x}_g^{(0)}
    \right\|_2
    <
    \frac{1}{\sqrt{2}}
    \quad\Longrightarrow\quad
    \mathbf{q}_g^{(t)}
    =
    \mathbf{c}_{i_g^{(0)}}.
    \label{eq:method_init_stability}
\end{equation}
To see this, write \(\mathbf{c}=\mathbf{c}_{i_g^{(0)}}\).
The triangle inequality gives \(\delta=\|\mathbf{x}_g^{(t)}-\mathbf{c}\|_2<1/\sqrt{2}\).
Every other lattice point \(\mathbf{c}'\) then satisfies
\[
    \|\mathbf{x}_g^{(t)}-\mathbf{c}'\|_2
    \geq \sqrt{2}-\delta
    > \delta,
\]
so \(\mathbf{c}\) remains the unique nearest point.

\section{Sparse Encoding and Decoding Verification}
\label{app:sparse_encoding}

\paragraph{Coordinate records.}
Each eight-weight group retains a 16-bit core index.
An escaped group additionally stores its 32-bit tensor-local position and doubled lattice coordinates,
\begin{equation}
    \mathbf{v}_g
    =
    2\mathbf{q}_g^\star
    \in\mathbb{Z}^{8}.
    \label{eq:method_escape_coordinates}
\end{equation}
Every \(E_8\) point has all-integer or all-half-integer coordinates, so \(\mathbf{v}_g\) is integral.
Eight signed 16-bit coordinates and one 32-bit position occupy 20 bytes per record.
During decoding, the record supplies the absolute lattice coordinates and overrides the fallback codeword, so the exported point is the lattice point selected in training.

Export checks that every position fits the selected index representation and that the stored coordinates satisfy
\[
    -2^{15}
    \leq (\mathbf{v}_g)_j
    \leq 2^{15}-1,
    \qquad j=1,\ldots,8.
\]

\paragraph{Conditional root packing.}
A smaller record is available when an escaped point differs from a selected core fallback by one \(E_8\) root.
Define
\begin{equation}
    \mathcal{R}
    =
    \left\{
        \mathbf{r}\in E_8:
        \|\mathbf{r}\|_2^2=2
    \right\},
    \qquad |\mathcal{R}|=240.
    \label{eq:method_root_set}
\end{equation}
When the residual belongs to \(\mathcal{R}\), decoding becomes
\begin{equation}
    \mathbf{q}_g^\star
    =
    \mathbf{c}_{i_g}+\mathbf{r}_{j_g},
    \qquad \mathbf{r}_{j_g}\in\mathcal{R}.
    \label{eq:method_root_decode}
\end{equation}
An 8-bit root identifier and a 24-bit position within a chunk form a 4-byte record, excluding chunk metadata.
All \(26{,}435\) escaped groups in the 375-step artifact satisfy this condition.
Payloads reported in the paper use the 20-byte coordinate record, including the 750-step endpoint.

\paragraph{Fused decoding.}
For the corresponding group of input-side transformed activations \(\mathbf{h}_g\), linearity gives
\begin{equation}
    \left\langle\mathbf{h}_g,\mathbf{q}_g^\star\right\rangle
    =
    \left\langle\mathbf{h}_g,\mathbf{c}_{i_g}\right\rangle
    +
    \left\langle\mathbf{h}_g,\mathbf{r}_{j_g}\right\rangle.
    \label{eq:method_fused_correction}
\end{equation}
The fused matrix-vector path adds the sparse correction to the core-codeword contribution before the output-side Hadamard transform.
The decoding scale \(c_s\) applies to both lattice terms, and the offset \(c_z\) is applied once.
The corrected weight matrix is never materialized.

\paragraph{Encoding checks.}
Table~\ref{tab:app:encoding_checks} records two checks.
Core decoding agrees on all \(65{,}536\) codewords, and every escaped group in the 375-step artifact lies one root from its fallback.
The second count is consistent with \(\rho=6.540\times10^{-6}\) over the \(4.04\times10^{9}\) quantized groups of the model.

\begin{table}[t]
    \centering
    \small
    \renewcommand{\arraystretch}{1.12}
    \caption{Encoding checks. Root packing holds for every escaped group in the 375-step artifact.}
    \label{tab:app:encoding_checks}
    \begin{tabular}{@{}p{0.28\linewidth}p{0.32\linewidth}p{0.32\linewidth}@{}}
        \toprule
        Check & Coverage & Result \\
        \midrule
        Core decoding agreement
        & All 65,536 core codewords
        & Identical lattice coordinates across the tested decoding paths \\
        Single-root membership
        & All 26,435 escaped groups in the 375-step artifact
        & Every residual belongs to \(\mathcal{R}\) \\
        \bottomrule
    \end{tabular}
\end{table}

\section{Additional Experimental Results}
\label{app:exp_details}

\subsection{Training and Evaluation Protocol}
\label{app:exp_protocol}

Table~\ref{tab:app:training_config} records the Qwen3.6-35B-A3B settings.
The quality comparison uses the 750-step endpoints.
The paired timing runs use an independently annealed 375-step schedule.
The Llama-3.1-8B second stage uses that same 375-step shape.

\begin{table}[t]
    \centering
    \small
    \renewcommand{\arraystretch}{1.08}
    \caption{Recorded settings for the Qwen3.6-35B-A3B experiments.}
    \label{tab:app:training_config}
    \begin{tabular}{@{}p{0.34\linewidth}p{0.61\linewidth}@{}}
        \toprule
        Setting & Value \\
        \midrule
        Model & Qwen3.6-35B-A3B \\
        Total parameters & \(35{,}107{,}181{,}936\) \\
        Quantized parameters
        & \(32{,}338{,}083{,}840\) expert FFN weights (\(92.1\%\)) \\
        Unquantized weights
        & \(2{,}769{,}098{,}096\) in BF16: attention projections, router, embeddings, norms \\
        Rate, quantized weights & 2.03 bits, including scales and codebooks \\
        Latent weights & BF16 \\
        Projection candidates and distances & FP32 \\
        PTQ preprocessing
        & LDLQ with block-Hadamard transforms of size 128 \\
        Finite core
        & \(2^{16}\) codewords; eight coordinates per group \\
        Training corpus
        & 2,000 long sequences from UltraData \citep{ultradata} \\
        Tokens per corpus pass & Approximately 50.8 million \\
        Maximum sequence length & 61,959 tokens \\
        Configured context window & 65,536 tokens \\
        Main quality schedule & 750 steps over six epochs \\
        Peak learning rate & \(10^{-4}\), warmup--stable--decay to zero \\
        Paired timing configuration & 375 steps, eight GPUs \\
        \bottomrule
    \end{tabular}
\end{table}

\paragraph{Comparison controls.}
E8-QAT and LC-QAT \citep{lcqat} share the base model, the quantized tensors, the training corpus, and the step budget.
Timing measurements cover the QAT stage.

\paragraph{Baselines and stages.}
E8-PTQ is the lattice checkpoint that initializes E8-QAT, and UniSVQ is the scalar checkpoint that initializes LC-QAT.
The matched-bit AQLM baseline uses one learned \(2^{16}\)-entry codebook shared by all feed-forward matrices of a layer.
QuIP\# and QTIP are included as additional vector PTQ baselines.
The per-matrix AQLM configuration, at \(7.36\) bits, is reported as a higher-bit reference.

\paragraph{Downstream evaluation.}
We evaluate function calling with BFCL \citep{bfcl}, scientific question answering with GPQA \citep{gpqa}, code generation with LiveCodeBench \citep{livecodebench}, and instruction following with IFEval \citep{ifeval}.
The reported metrics are BFCL v3 overall accuracy, GPQA accuracy, LiveCodeBench pass@1, and IFEval prompt-level strict accuracy.
All models use the same harness and the same generation settings.

\subsection{Stage-1 Reconstruction Quality}
\label{app:stage1}

We evaluate E8-PTQ, UniSVQ, and layer-shared AQLM with the same calibration Hessians, collected once on the original model and reused for every method.
Per-expert Hessians are shrunk toward the layer mean with a pseudo-count equal to the input width.
For layer \(\ell\), with feed-forward matrices \(\mathcal{M}_{\ell}\) and routed-token weights \(w_{\ell m}\),
\begin{equation}
    D_{\ell}
    =
    \sum_{m\in\mathcal{M}_{\ell}}
    w_{\ell m}
    \frac{
        \operatorname{tr}\!\left(
            \Delta\mathbf{W}_{\ell m}
            H_{\ell m}
            \Delta\mathbf{W}_{\ell m}^{\top}
        \right)
    }{
        \operatorname{tr}\!\left(
            \mathbf{W}_{\ell m}
            H_{\ell m}
            \mathbf{W}_{\ell m}^{\top}
        \right)
    }.
    \label{eq:app:stage1_distortion}
\end{equation}
Shared-expert matrices use unit weight.
The reported mean averages \(D_{\ell}\) over the 40 decoder layers.

E8-PTQ has the lowest distortion in 39 of the 40 layers.
Its average, \(0.0482\), is \(20.4\%\) below UniSVQ and \(19.4\%\) below layer-shared AQLM.

\begin{table}[t]
    \centering
    \small
    \renewcommand{\arraystretch}{1.08}
    \caption{Stage-1 reconstruction on Qwen3.6-35B-A3B. Bold marks the lowest distortion and the largest best-layer count.}
    \label{tab:app:stage1_distortion}
    \begin{tabular*}{\linewidth}{@{\extracolsep{\fill}}lrr@{}}
        \toprule
        Method & Mean distortion \(\downarrow\) & Best layers \(\uparrow\) \\
        \midrule
        UniSVQ & 0.0605 & 0/40 \\
        AQLM, layer-shared & 0.0598 & 1/40 \\
        E8-PTQ & \textbf{0.0482} & \textbf{39/40} \\
        \bottomrule
    \end{tabular*}
\end{table}

\subsection{Additional Quality Results}
\label{app:full}
\label{app:other_metrics}

\paragraph{Complete task breakdown.}
Table~\ref{tab:app:lmeval} adds ARC-Easy and BoolQ to the four tasks in Table~\ref{tab:exp:main}.
ARC-Challenge, ARC-Easy, PIQA, and HellaSwag use normalized accuracy; WinoGrande and BoolQ use accuracy.
E8-QAT obtains the highest six-task average among the 2-bit configurations and leads on five of the six tasks.
On BoolQ, UniSVQ scores \(89.33\).

\begin{table}[t]
    \centering
    \small
    \setlength{\tabcolsep}{3pt}
    \renewcommand{\arraystretch}{1.08}
    \caption{Six-task \texttt{lm\_eval} results on Qwen3.6-35B-A3B. Bold marks the best 2-bit score in each column.}
    \label{tab:app:lmeval}
    \begin{tabular*}{\linewidth}{@{\extracolsep{\fill}}llrrrrrrrr@{}}
        \toprule
        Stage & Method & Bits
        & ARC-C & ARC-E & HellaSwag
        & PIQA & WinoGrande & BoolQ & Avg\(_6\) \\
        \midrule
        Ref. & BF16 & 16
        & 55.20 & 72.22 & 83.49
        & 82.75 & 74.27 & 85.02 & 75.49 \\
        \midrule
        PTQ & UniSVQ & 2.03
        & 56.83 & 75.84 & 78.31
        & 80.58 & 71.59 & \textbf{89.33} & 75.41 \\
        PTQ & AQLM, layer-shared & 2.03
        & 54.95 & 73.06 & 79.59
        & 81.50 & 73.32 & 88.38 & 75.13 \\
        PTQ & E8-PTQ & 2.03
        & 56.74 & 78.41 & 79.38
        & 80.47 & 72.61 & 88.78 & 76.06 \\
        \midrule
        QAT & LC-QAT & 2.03
        & 56.57 & 75.72 & 81.28
        & 80.74 & 73.88 & 87.61 & 75.97 \\
        QAT & E8-QAT & 2.03
        & \textbf{57.85} & \textbf{79.34} & \textbf{82.29}
        & \textbf{81.56} & \textbf{74.03}
        & 86.64 & \textbf{76.95} \\
        \midrule
        \multicolumn{10}{@{}l}{\textit{Higher-bit reference}} \\
        PTQ & AQLM, per-matrix & 7.36
        & 56.91 & 74.87 & 82.18
        & 81.77 & 75.14 & 87.61 & 76.42 \\
        \bottomrule
    \end{tabular*}
\end{table}

\paragraph{Additional reasoning evaluation.}
On AIME25 pass@3, E8-QAT matches BF16 and LC-QAT at \(90.00\).
AQLM scores \(83.33\).

\begin{table}[t]
    \centering
    \small
    \renewcommand{\arraystretch}{1.08}
    \caption{AIME25 pass@3. Bold marks the highest score, including ties.}
    \label{tab:app:aime25}
    \begin{tabular}{@{}lr@{}}
        \toprule
        Method & AIME25 pass@3 \(\uparrow\) \\
        \midrule
        BF16 & \textbf{90.00} \\
        AQLM & 83.33 \\
        LC-QAT & \textbf{90.00} \\
        E8-QAT & \textbf{90.00} \\
        \bottomrule
    \end{tabular}
\end{table}

\subsection{Llama-3.1-8B Results}
\label{app:llama}

Table~\ref{tab:app:llama} adds ARC-Easy, BoolQ, and WikiText perplexity to Table~\ref{tab:exp:llama}.
E8-QAT has the highest six-task average and the lowest perplexity among the quantized models: \(13.77\), against \(16.73\) for LC-QAT and \(9.01\) for BF16.

The last row uses the Qwen peak learning rate, \(10^{-4}\), on Llama.
The Llama feed-forward scale \(w_s\) is \(2.68\times\) larger, so this rate pushes \(0.607\%\) of groups outside the finite core and raises the index-and-extension payload to \(2.12\) bits per quantized weight.
Scaling the rate by the inverse weight scale, to \(3.73\times10^{-5}\), is the E8-QAT result in Table~\ref{tab:exp:llama}.

\begin{table}[t]
    \centering
    \small
    \setlength{\tabcolsep}{3pt}
    \renewcommand{\arraystretch}{1.08}
    \caption{Complete \texttt{lm\_eval} results on Llama-3.1-8B. WikiText is perplexity (lower is better). Bold marks the best quantized result; the last row is excluded.}
    \label{tab:app:llama}
    \begin{tabular*}{\linewidth}{@{\extracolsep{\fill}}llrrrrrrrr@{}}
        \toprule
        Stage & Method & ARC-C & ARC-E & HelS & PIQA & WinG & BoolQ
        & Avg\(_6\) & WikiText \\
        \midrule
        Ref. & BF16
        & 55.55 & 79.80 & 79.55 & 81.28 & 73.80 & 85.47 & 75.91 & 9.01 \\
        \midrule
        PTQ & UniSVQ
        & 39.51 & 65.36 & 63.55 & 72.63 & 66.46 & 79.27 & 64.46 & 22.38 \\
        PTQ & E8-PTQ
        & 43.94 & \textbf{70.50} & 66.16 & 74.76 & 68.90 & 81.47 & 67.62
        & 20.79 \\
        \midrule
        QAT & LC-QAT
        & 40.10 & 64.14 & 67.71 & 73.29 & 67.40 & \textbf{82.57} & 65.87
        & 16.73 \\
        QAT & E8-QAT
        & \textbf{44.62} & 69.49 & \textbf{70.03} & \textbf{76.17}
        & \textbf{69.61} & 77.43 & \textbf{67.89} & \textbf{13.77} \\
        \midrule
        QAT & E8-QAT, LR \(10^{-4}\)
        & 35.84 & 52.36 & 58.39 & 69.48 & 59.04 & 72.29 & 57.90 & 19.46 \\
        \bottomrule
    \end{tabular*}
\end{table}

\subsection{Escape Measurements and Learning-Rate Sensitivity}
\label{app:escape_growth}

\paragraph{Escape across training budgets.}
Table~\ref{tab:escape-trajectories} reports the fraction \(\rho\) of groups outside \(\mathcal{C}_{\mathrm{core}}\) for the 375-step and 750-step schedules.
Escapes stay sparse, and growth slows near the endpoints: the final 50 steps of the 750-step run raise \(\rho\) by \(0.25\%\) of its own value.

\begin{table}[t]
    \centering
    \small
    \renewcommand{\arraystretch}{1.05}
    \caption{Escape fractions under the two annealing schedules. A zero is an exact zero.}
    \label{tab:escape-trajectories}
    \begin{tabular}{@{}rrr@{}}
        \toprule
        Training budget & Checkpoint step & Escape fraction \(\rho\) \\
        \midrule
        375 & 100 & \(0\) \\
        375 & 200 & \(1.196\times10^{-6}\) \\
        375 & 300 & \(5.672\times10^{-6}\) \\
        375 & 375 & \(6.540\times10^{-6}\) \\
        \midrule
        750 & 400 & \(1.618\times10^{-5}\) \\
        750 & 500 & \(3.618\times10^{-5}\) \\
        750 & 600 & \(7.077\times10^{-5}\) \\
        750 & 700 & \(8.535\times10^{-5}\) \\
        750 & 750 & \(8.556\times10^{-5}\) \\
        \bottomrule
    \end{tabular}
\end{table}

\paragraph{Learning-rate sensitivity.}
Two 375-step runs share the schedule shape \(11+252+112\) and differ only in the peak learning rate.
Halving it from \(10^{-4}\) to \(5\times10^{-5}\) lowers the endpoint escape fraction by \(40.7\times\), from \(6.540\times10^{-6}\) to \(1.608\times10^{-7}\).
The reduced-rate run has zero escapes at steps 100 and 200.

\paragraph{Storage consequences.}
Each group keeps a 16-bit core index, and each escaped group adds a 20-byte coordinate record:
\begin{equation}
    \Delta b = 20\rho,
    \qquad
    b_{\mathrm{payload}} = 2+\Delta b.
    \label{eq:app:escape_payload}
\end{equation}
At the 750-step endpoint the sparse records add \(1.711\times10^{-3}\) bits per quantized weight, \(0.086\%\) of the 2-bit index payload.
These figures cover indices and escape records.

\begin{table}[t]
    \centering
    \small
    \setlength{\tabcolsep}{4pt}
    \renewcommand{\arraystretch}{1.08}
    \caption{Endpoint storage with 20-byte coordinate records, in bits per quantized weight.}
    \label{tab:app:escape_payload}
    \begin{tabular*}{\linewidth}{@{\extracolsep{\fill}}lrrr@{}}
        \toprule
        Endpoint & \(\rho\) & \(\Delta b\) & \(b_{\mathrm{payload}}\) \\
        \midrule
        375 steps, peak LR \(10^{-4}\)
        & \(6.540\times10^{-6}\)
        & \(1.308\times10^{-4}\)
        & \(2.000131\) \\
        375 steps, peak LR \(5\times10^{-5}\)
        & \(1.608\times10^{-7}\)
        & \(3.216\times10^{-6}\)
        & \(2.000003\) \\
        750 steps
        & \(8.556\times10^{-5}\)
        & \(1.711\times10^{-3}\)
        & \(2.001711\) \\
        \bottomrule
    \end{tabular*}
\end{table}

\section{Limitations}
\label{app:limitations}

\paragraph{Evaluation scope.}
Our experiments cover Qwen3.6-35B-A3B and Llama-3.1-8B, with long-form generation and agentic evaluations focused on the MoE model.
Extending the evaluation to additional architectures, model scales, and application domains would further characterize the applicability of the method.

\paragraph{Weight-only quantization.}
This work quantizes weights.
Activations and the KV cache remain in higher precision.
Combining the lattice weights with activation and KV-cache quantization is a direct way to reduce memory further in long-context inference.

\paragraph{Deployment configurations.}
The inference results use the fused kernels and batch sizes reported in the main text.
The memory reduction is what admits the larger batch.
Further kernel work and other hardware platforms are natural next measurements.

\end{document}